\documentclass[a4paper,conference]{article}

\usepackage[pdftex]{graphicx}
\graphicspath{{./images}}
\DeclareGraphicsExtensions{.pdf,.jpeg,.png}

\usepackage{url}
\usepackage{hyperref}
\hypersetup{
    colorlinks=true,
    urlcolor=magenta
}

\usepackage{multirow}
\usepackage[table,xcdraw]{xcolor}

\begin{document}
%
\title{On Embodied Visual Navigation\\ in Real Environments Through Habitat\thanks{Published in International Conference on Pattern Recognition (ICPR), 2020.}}
\date{}

\author{
Marco Rosano\textsuperscript{1,3},
Antonino Furnari\textsuperscript{1},\\
Luigi Gulino\textsuperscript{3},
Giovanni Maria Farinella\textsuperscript{1,2}\\\\
\textsuperscript{1}FPV@IPLAB - Department of Mathematics and Computer \\Science, University of Catania, Italy\\
\textsuperscript{2}Cognitive Robotics and Social Sensing Laboratory,\\
ICAR-CNR, Palermo, Italy,\\
\textsuperscript{3}OrangeDev s.r.l., Firenze, Italy\\\\
marco.rosano@unict.it, furnari@dmi.unict.it, \\
luigi.gulino@orangedev.it,  gfarinella@dmi.unict.it
}


%


\maketitle
\begin{abstract}
Visual navigation models based on deep learning can learn effective policies when trained on large amounts of visual observations through reinforcement learning. 
Unfortunately, collecting the required experience in the real world requires the deployment of a robotic platform, which is expensive and time-consuming. 
To deal with this limitation, several simulation platforms have been proposed in order to train visual navigation policies on virtual environments efficiently. 
Despite the advantages they offer, simulators present a limited realism in terms of appearance and physical dynamics, leading to navigation policies that do not generalize in the real world. 
In this paper, we propose a tool based on the Habitat simulator which exploits real world images of the environment, together with sensor and actuator noise models, to produce more realistic navigation episodes. 
We perform a range of experiments to assess the ability of such policies to generalize using virtual and real-world images, as well as observations transformed with unsupervised domain adaptation approaches. 
We also assess the impact of sensor and actuation noise on the navigation performance and investigate whether it allows to learn more robust navigation policies.
We show that our tool can effectively help to train and evaluate navigation policies on real-world observations without running navigation episodes in the real world.
\end{abstract}


%

\section{Introduction}

\label{sec:intro}

The autonomous visual navigation problem has been studied for years by the research community and has recently attracted even more interest given its potential range of applications in real contexts~\cite{bonin2008visual}.
Recently, Deep Learning (DL) approaches have shown that it is possible to learn a navigation policy in an end-to-end fashion directly from visual observations collected by the agent while interacting with the environment~\cite{zhu2017target,gupta_cognitive}.
This approach outperforms past navigation paradigms and can learn a more abstract representation of the environment which allows to easily transfer the acquired knowledge to unseen contexts~\cite{habitat19iccv,chaplot2020learning}.
One of the major drawbacks of DL-based navigation approaches is that they require to run several navigation episodes in order to learn effective navigation policies. 
Unfortunately, collecting the required experience in the real-world is difficult due to the maintenance cost required by a robotic platform acting in a trial-and-error setup, the variety of the scenes to be explored to ensure regularization, as well as the time needed to perform all the training episodes.
\begin{figure}[t]
    \centering
    \includegraphics[width=0.99\linewidth]{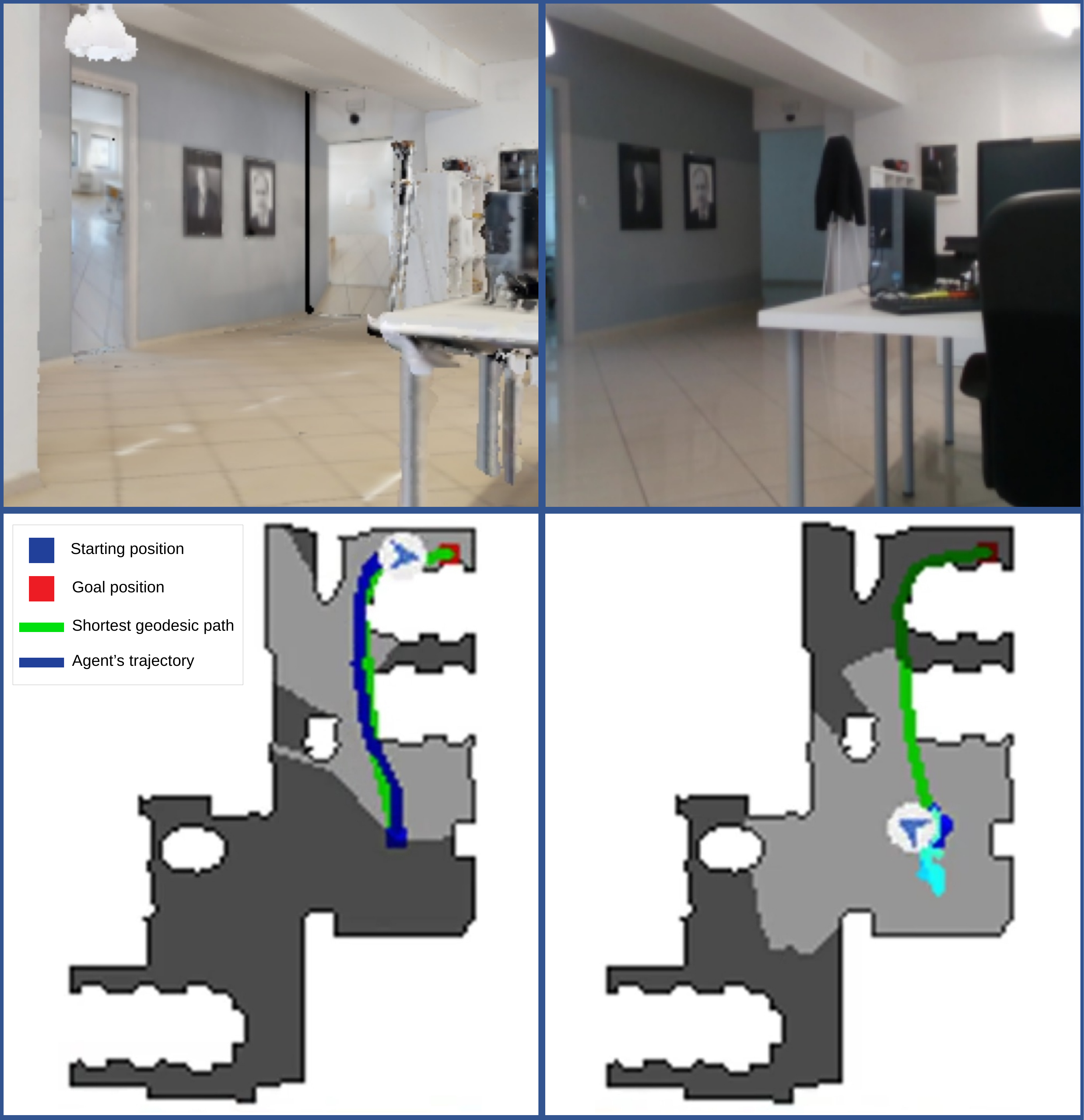}
    \caption{Navigation examples performed by an agent trained in a virtual environment when tested on (left) virtual observations and (right) real-world observations. The top row shows examples of observations seen during test. The policy in the virtual environment does not generalize to real-world observations.}
    \label{figure: nav_trajectories}
\end{figure}
A popular approach to overcome these limitations consists in collecting all the required experience in simulated environments which replicate the appearance of real scenes.
Following this idea, several simulators for training embodied agents have been proposed in previous works~\cite{habitat19iccv,savva2017minos,xiazamirhe2018gibsonenv}.
These simulators allow to import 3D models of realistic environments previously acquired with a 3D scanner, resulting in a fast increase of the available training environments.

Even if training navigation agents with these tools has been shown to be effective, transferring the learned policies to the real world (e.g. to a real robotic platform) is still a challenging and open problem. 
This is mainly due to a domain shift caused by a number of factors such as the difference in the appearance between real-world visual observations and the ones provided by the simulator, the absence of real-world noise in the simulation (i.e., actuator noise and sensor noise), as well as the simplified physical interactions with the elements of the scene (e.g. no collisions, no friction, no uneven ground, etc.).
Indeed, a navigation policy learned in a simulated environment performs very poorly when transferred to a real robotic agent who has to explore the environment~\cite{robothor}. Figure~\ref{figure: nav_trajectories} shows two navigation examples obtained by an agent trained with virtual observations. When the agent is tested on visual observations coming from the simulator (Fig.~\ref{figure: nav_trajectories} left), the navigation is close to perfect, whereas when the same agent is tested with real-world images (Fig.~\ref{figure: nav_trajectories} right), the navigation fails.
\begin{figure}[!t]
    \centering
    \includegraphics[width=\linewidth]{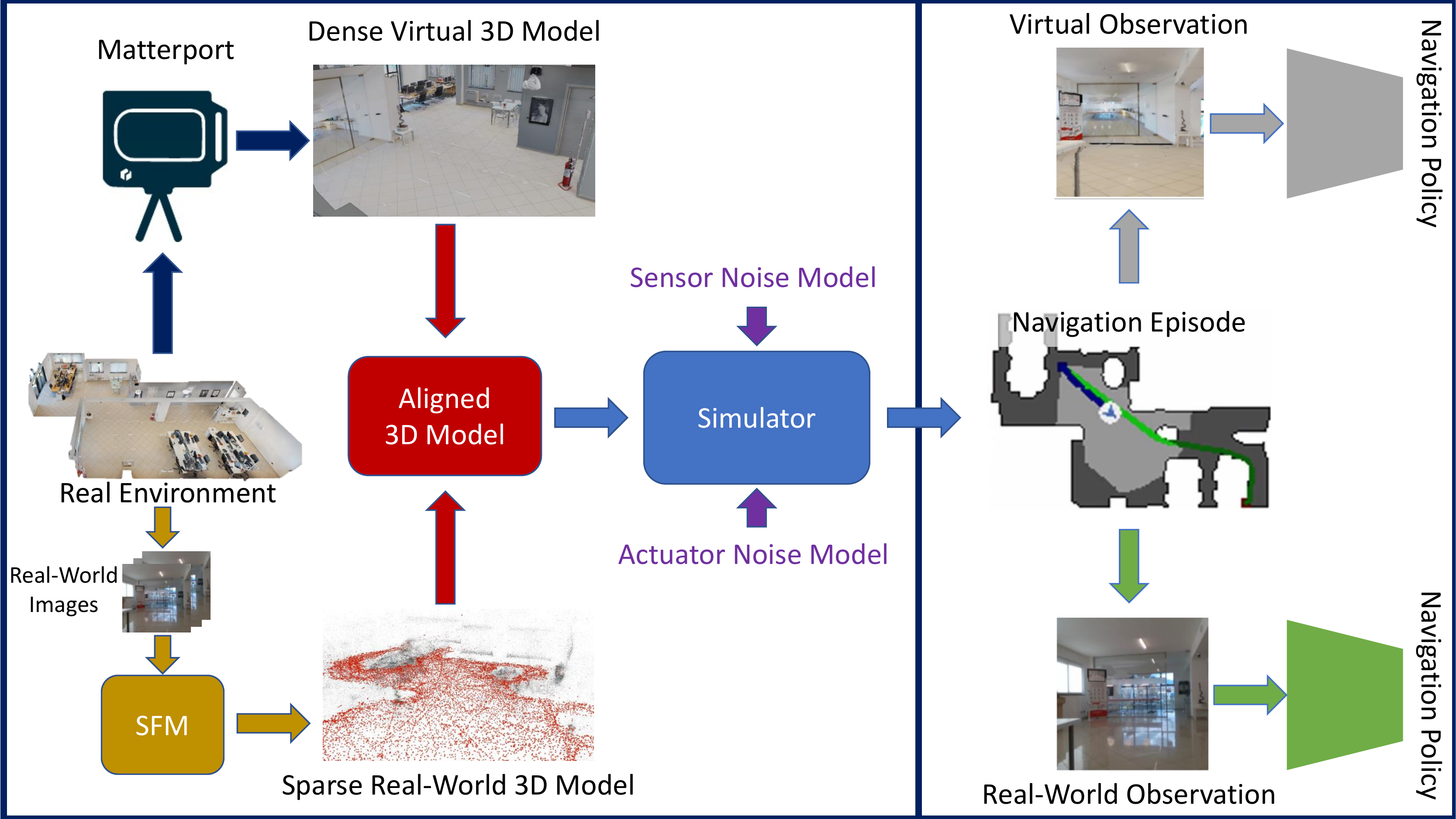}
    \caption{Our method exploits two 3D models of the same environment to generate realistic navigation episodes comprising both virtual and real-world observations, which can be used to evaluate or train navigation policies. The model can also account for sensor and actuator noise.}
    \label{figure: pipeline}
\end{figure}
In this paper, we investigate how a navigation policy learned in a simulated world transfers to a corresponding real environment.
Previous works have tried to measure and compensate for this domain shift by fine-tuning models learned in simulation using few real observations sampled on a sparse grid~\cite{zhu2017target}, or by running parallel experiments in simulation and in reality~\cite{robothor,arewemakingprogress}. While these approaches are viable, they considered simplified virtual environments~\cite{zhu2017target} or require access to a real robotic platform during training and test~\cite{robothor,arewemakingprogress}.
We argue that this prevents the exploitation of the convenient abstraction provided by a simulator.
Hence, we propose to perform the analysis by relying on a simulator to sample real-world observations, which allows to take advantage of the fast and inexpensive training and evaluation provided by modern simulators.
The core of our approach is a tool based on the popular Habitat simulator~\cite{habitat19iccv} which allows to generate navigation episodes comprising virtual and real-world observations. The scheme of the approach is illustrated in \figurename~\ref{figure: pipeline}.
Given a real environment, we construct two 3D models: 1) a ``virtual'' model using the Matterport 3D scanner\footnote{https://matterport.com/}, and 2) a ``real-world'' model running a Structure from Motion algorithm (SfM)~\cite{schoenberger2016sfm} on real images of the environment.
The ``virtual'' model is an accurate representation of the world with limited photo-realism, whereas the ``real-world'' model is a sparse collection of real images attached with their related camera pose.
The two models are then geometrically registered so that their coordinate systems match to form an ``aligned 3D model''.
The proposed tool hence allows to generate paired virtual and real-world navigation episodes by sampling visual observations from the aligned 3D model.
Sensor and actuator noise models can be provided to enable the simulator to generate more realistic navigation episodes. 
The developed tool allows to 1) assess the performance of a navigation model which has been trained with virtual observations, when tested on real-world visual inputs; 2) measure the impact of sensors noise, actuator noise and appearance gap on a navigation policy learned with virtual observations when tested on real-world visual inputs; 3) adapt a navigation model learned with virtual observations to the real domain by fine-tuning it with real-world observations; 4) train realistic navigation policies introducing simulated actuator and sensor noise akin to the one which can be observed in real robotic platforms.

We perform experiments on an office environment of $150m^2$. Our analysis points out that navigation policies trained with virtual observations perform poorly when tested in real-world scenes, noisy sensors and noisy actuators have a significant impact on the navigation performance, models trained in a noisy environment can learn navigation policies which are more robust to noise.
We also benchmark different approaches to improve the performance of navigation agents when deployed to the real world. We found that both unsupervised domain adaptation such as CycleGAN~\cite{cyclegan} and supervised adaptation such as fine-tuning, allow to adapt the navigation policy to the real-world scenario, with the latter class of approaches obtaining better performance.
Results suggest that, while the approaches to domain adaptation for visual navigation are promising, much work is still to be done to achieve results which are exploitable in the real world. We hope that the proposed tool can foster research in this direction.

In sum, the contributions of this work are as follows:
\begin{enumerate}
    \item we propose a tool based on Habitat~\cite{habitat19iccv} to train and evaluate entirely in simulation visual navigation policies on real observations and with realistic sensor and actuator noise;
    \item we show that a navigation model trained purely on virtual observations performs poorly when tested on the corresponding real environment and how domain adaptation approaches can help to transfer the learned policy to the real world;
    \item we investigate the impact of sensor and actuator noise models on navigation performance by comparing navigation models trained with and without noise and tested in a noisy environment. 
\end{enumerate}

\section{Related Work}

\label{sec:related}

Previous works have considered the problem of assessing the transfer of a navigation policy from a virtual environment to a real one. Specifically, The authors of~\cite{zhu2017target} proposed to fine-tune navigation agents on a grid of real-world images acquired from specific locations. Differently from this approach, we rely on a much denser collection of real-world images to train and test our models, which is possible thanks to the proposed tool. The authors of~\cite{robothor, arewemakingprogress} recently proposed to perform paired experiments in simulation and reality. However, this requires access to a robotic platform in the training and testing stages. Our approach aims to move the analysis as much as possible to simulation. While research on this specific topic is still emerging, our work is related to other lines of research including embodied navigation simulators, visual navigation, and virtual-to-real domain adaptation. 

\subsection{Embodied Navigation Simulators}
The development of advanced simulators~\cite{savva2017minos,xiazamirhe2018gibsonenv,robothor,house3d,kolve2017ai2,habitat19iccv} together with the acquisition of large-scale virtual indoor environments ~\cite{replica19arxiv,Matterport3D,xiazamirhe2018gibsonenv,song2016ssc} have been instrumental to speed up the research on autonomous agents and give them the ability to navigate environments and interact with humans.
These simulators allow to train navigation algorithms based on deep learning which would be unfeasible to train directly in real-world environments~\cite{gupta_cognitive,chaplot2020learning,chen2018learning,Wijmans2020DD-PPO:}.

Despite such advances, transferring a navigation policy trained in simulation directly to real contexts is not straightforward and still a problem under exploration~\cite{zhu2017target,robothor,arewemakingprogress}.
Our work builds on previous work on simulators for visual navigation~\cite{habitat19iccv}, aiming to augment them with realistic noise models and real-world observations.

\subsection{Visual Navigation}

In end-to-end visual navigation, the policy is learned by the agent given egocentric observations and the goal to reach. 
If the goal is given as the coordinates of the target or as an image, the task is referred to as \textit{geometric navigation}. 
In this case, the agent is encouraged to learn the geometry of the environment in order to complete the task. 
Previous works have addressed geometric navigation in different settings, such as providing images of the goal as target for the agent~\cite{zhu2017target}, jointly learning the mapping of the environment and the policy to reach the destination in a supervised way~\cite{gupta_cognitive}, explicitly modelling environment mapping through SLAM and both long- and short-term goal estimation~\cite{chaplot2020learning}.
Other authors have shown that it is possible to exploit the efficiency of the Habitat simulator~\cite{habitat19iccv} to train a mapless PointGoal navigation policy with millions~\cite{anderson2018evaluation} or even billions~\cite{Wijmans2020DD-PPO:} of frames, greatly outperforming classic approaches.

In this work, we consider the settings of geometric navigation in which the goal is provided in the form of coordinates relative to the agent's position~\cite{anderson2018evaluation}. 
In our investigation, we focus on the agent's ability to transfer the geometric representation learned in a virtual environment to its real counterpart.

\subsection{Virtual-to-Real Domain Adaptation}
Domain adaptation techniques can be employed to reduce the domain gap between virtual and real observations by learning domain-invariant image representations which allow to transfer action policies learned in simulation to the real world. 
Among the most notable approaches, domain randomization methods aim to improve generalization to real-world observations by increasing the size and variety of training data in tasks such as robot grasping~\cite{dom_rand_tobin} and visual navigation~\cite{sadeghi2016cad2rl,Loquercio2020DeepDR}.
Other approaches focus on learning domain invariant task-specific~\cite{zhang2017curriculum} or task-agnostic~\cite{sun2016return} higher order statistics of the scenes which are then used to train the agent's policy.
Recent works have used Generative Adversarial Networks (GANs)~\cite{gan} to transform the appearance of images from a domain to another~\cite{cyclegan,rao2020rl, hoffman2018cycada}.

In this work, we assess the impact of simple adaptation techniques on the ability to transfer a policy learned with simulated observations to real-world observations.
Specifically, we consider fine-tuning on real observations, which is made possible thanks to the proposed framework, and adapting virtual observations with CycleGAN~\cite{cyclegan}, which is an unsupervised domain adaptation approach as it does not require labeled real images.

\section{Embodied Visual Navigation In Real Environments Through Habitat}

\label{sec:proposed}

The proposed tool is based on the popular Habitat simulator~\cite{habitat19iccv}, an open-source embodied AI 3D platform provided with flexible APIs that allow researchers to access and extend its functionalities. Habitat can render RGB, depth and semantic images at up to $10,000$ fps and allows to import a variety of scenes, including custom environments acquired with a 3D scanner such as Matterport 3D.
Our tool, as well as the dataset composed by 3D models and images used for evaluation purposes are publicly available at the following link: \url{https://iplab.dmi.unict.it/EmbodiedVN/}.
As illustrated in Figure~\ref{figure: pipeline} and briefly discussed in the introduction, the proposed approach involves the acquisition of two 3D models of the environment, the alignment of the models, and the generation of navigation episodes using virtual and real-world observations, considering also sensor and actuator noise. We give details for each of these steps in the following sections.

\subsection{Acquisition and Alignment of the 3D Models}
As briefly mentioned in the introduction, the ``virtual'' 3D model is a geometrically accurate representation of the environment that can be imported in a simulator to perform efficient navigation episodes. 
The virtual model can be acquired using a scanner such as Matterport 3D, which is able to collect RGB and depth observations from different regions of the environment to reconstruct the 3D mesh of the space.
The main limitation of the resulting 3D model is that its photo-realism is limited. The top part of Figure~\ref{figure: nav_trajectories} compares a visual observation sampled from a virtual 3D model (on the left) to a corresponding real-world image (on the right).

\begin{figure}[!t]
    \centering
    \includegraphics[width=0.99\linewidth]{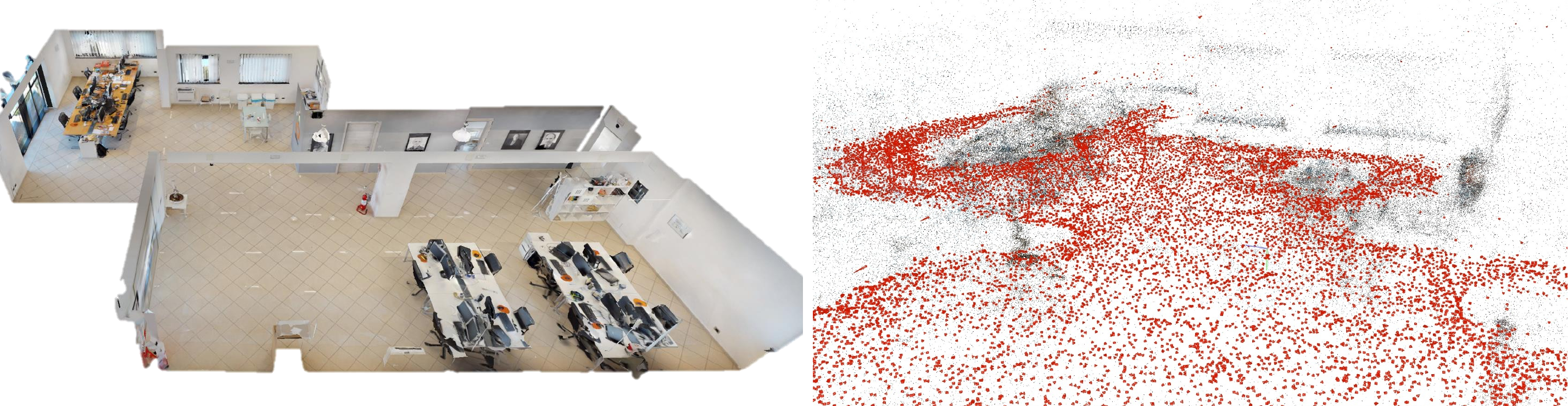}
    \caption{A view of the virtual (left) and real-world (right) 3D models. The virtual model is geometrically accurate and allows to sample images from any position but it lacks in photo-realism. The real-world model is a sparser collection of localized real-world images (each red marker represents the position of an image)}
    \label{figure: virtual_real_images}
\end{figure}

The ``real-world'' 3D model consists in a sparse reconstruction of the environment which does not provide an accurate geometric representation of the scene but contains photo-realistic images. 
This model can be constructed starting from real images collected from the environment using a digital camera or a real robotic platform (e.g. with random navigation). 
The set of sparse RGB images can be used to obtain a sparse 3D point cloud of the environment in which each image is labeled with its camera pose, using a Structure from Motion (SfM) algorithm~\cite{schoenberger2016sfm}\footnote{We use the COLMAP software: \url{https://colmap.github.io/faq.html}.}. Figure \ref{figure: virtual_real_images} compares the two models.
We would like to note that the virtual 3D model can be loaded into the Habitat simulator using the official API, whereas the real-world 3D model cannot be easily interfaced with Habitat without the proposed tool.

Since the virtual and the real 3D models are constructed independently, they are characterized by different coordinate systems. We align them using an image-based registration procedure that allows to transform the coordinate system of the real-world 3D model to match the one of a reference set of images sampled from the same environment and annotated with their camera pose\footnote{We use the \textit{model\_aligner} function of the COLMAP software~\url{https://colmap.github.io/faq.html}}.
To obtain this reference set, we use the Habitat simulator to collect virtual images with random poses from the virtual 3D model. Despite the fact that the photo-realism of the virtual model is limited, we found the alignment procedure to succeed.

\subsection{Generating Navigation Episodes with Real-World Observations}
Once the virtual and the real 3D models are aligned, the proposed tool can produce navigation episodes with real-world observations. This is done by running the simulator on the virtual environment and systematically replacing virtual observations with the real-world which are closest in space to the current agent position. 
Specifically, at each navigation step, the current pose of the agent is used to perform a fast  retrieval of the nearest image in the real-world 3D model. 
The retrieved real image can be fed by the visual navigation model in the simulator in order to choose the action to perform, together with the information about the goal to reach and other possible signals useful for the task. The retrieval process is repeated at each time-step until the navigation episode is completed.

To simplify retrieval, we transformed the 6DoF camera pose to 3DoF coordinates, where the first two degrees of freedom represent the X and the Z coordinates (parallel to the ground plane), whereas the third degree of freedom represents its rotation angle with respect to the Y axis perpendicular to the XZ plane. 
The transformation to 3DoF is performed because we assume that the agent's altitude is constant during the navigation and that the agent's camera does not perform rotations along its pitch and roll angles.
To be able to retrieve angles using the cosine distance, we represent the camera rotation $\theta$ as a unit vector $(u,v) = (\cos \theta, \sin \theta)$.
We use the FAISS library~\cite{FAISS} to perform an efficient two-steps retrieval procedure. In the first step, we filter images by angle, thesholding the cosine similarity score between the query and target angles. This allows us to select cameras which look in the same direction as the query pose.
In the second step, we choose the image with the closest X-Z position among the filtered ones. 
In our experiments, we found that a cosine similarity threshold of $0.96$ allows to obtain reasonable results.

\subsection{Sensor and Actuator Noise Models}
One of the major differences between simulation and reality is the presence of noise in perception and actions. 
As emerged in our experiments, navigation models trained in simulation assuming a perfect odometry perform poorly when deployed in noisy contexts and the problem can be alleviated by training them with simulated noise. 
We implemented two noise models: one for the sensor module (perception module), the other for the actuation module (action module).
In our setup, the only sensor considered is a localization sensor which is used to determine the position of the agent at each time-step.
Our model can work with generic noise models that can be tuned and adapted to any specific robotic platform.
In this paper, we assume Gaussian noise for sensors and actuators.

\section{Experimental Setup}
\label{sec:exp}
In this section, we report details about the dataset acquisition, the simulation of the episodes, training and evaluation.

\subsection{Environment and Simulation}
For our experiments, we considered an office of about 150 square meters.
We reconstructed the virtual and real-world 3D models of the same environment as described in Section~\ref{sec:proposed}.
Specifically, the virtual 3D model has been acquired using Matterport 3D and imported in the Habitat simulator~\cite{habitat19iccv} to perform the virtual navigation episodes.
The real 3D model has been created by collecting $24,896$ RGB images using the Sanbot Elf\footnote{\url{http://en.sanbot.com/}} robotic platform equipped with an Intel RealSense~D435 camera\footnote{\url{https://www.intelrealsense.com/}}, following a simple random exploration procedure aimed at covering the whole space.
The collected images have been used to reconstruct the 3D space with COLMAP~\cite{schoenberger2016sfm}. With the reconstruction, we obtained as output the 6DoF camera pose of all images, relative to the reconstructed space.
The set of real-world images has been split into training and test sets of equal size, following a sampling strategy to ensure a uniform distribution of camera poses in both sets and that neighboring frames do not fall into different sets to avoid over-fitting.
We used the training set to train and fine-tune the navigation models on real-world images and the test set to evaluate their performances.

To align the 3D models, we used the aforementioned image-based alignment procedure offered by COLMAP~\cite{schoenberger2016sfm}, relying on a reference set of 6K images, labeled with their X, Y, Z camera coordinates. The images used for the alignment have been randomly sampled from the virtual environment using the Habitat simulator~\cite{habitat19iccv}.

We trained our models following the setup proposed in~\cite{Wijmans2020DD-PPO:}, performing our navigation episodes in the Habitat simulator~\cite{habitat19iccv}. The navigation is formulated as a PointGoal navigation task~\cite{anderson2018evaluation}. At each time-step, the agent receives the current RGB observation and the updated goal coordinates relative to the current agent's pose, then it chooses one of the four discrete actions to perform: \textit{move straight by} $0.25m$, \textit{turn left by} $10^\circ$, \textit{turn right by} $10^\circ$, \textit{STOP}.
The navigation episode ends when the \textit{STOP} action is performed or when the maximum number of execution steps is reached. The maximum number of steps is set to 200 in our experiments.

We use the RGB DD-PPO Reinforcement Learning model~\cite{Wijmans2020DD-PPO:} pre-trained on the Gibson~\cite{xiazamirhe2018gibsonenv} and Matterport~3D~\cite{Matterport3D} datasets.
The model is composed by a visual encoder and two recurrent layers. The visual encoder takes a $256\times256$ RGB image as input and outputs an embedding, which is concatenated with the information about the goal coordinates and the previous action performed. The resulting vector is then fed to a recurrent network which outputs the action to be performed.

\begin{table}[t]
\footnotesize{}
\caption{Noise levels considered in our experiments as standard deviations of our Gaussian noise models.}
\centering 
\begin{tabular}{c c c c}
\hline
\rule{0pt}{3ex}
 &  & Noise level &\\ [1ex]
\rule{0pt}{3ex}
 & Small & Medium & Large \\ [0.5ex]
\hline                  
\rule{0pt}{3ex}
Localization noise & $0.20m$; $7^\circ$ & $0.40m$; $15^\circ$  & $0.80m$; $30^\circ$ 
\\
\rule{0pt}{3ex}
Actuation noise & $0.05m$; $5^\circ$ & $0.10m$; $10^\circ$  & $0.20m$; $20^\circ$
\\ [0.5ex]
\hline
\end{tabular}
\label{table:noise_levels}
\end{table}

\subsection{Domain Adaptation of the Navigation Policy to the Real Environment}
We performed different experiments to evaluate the ability of the navigation model to adapt to real-world episodes. 
Specifically, we measured the performance of the navigation model tested on real-world images when:
\begin{enumerate}
    \item trained on virtual observations;
    \item trained on real observations;
    \item trained on virtual observations and fine-tuned on real observations;
    \item trained on virtual observations adapted with CycleGAN~\cite{cyclegan};
    \item trained on virtual observations and fine-tuned on virtual observations adapted with CycleGAN~\cite{cyclegan};
    \item trained on virtual observations adapted with CycleGAN~\cite{cyclegan} and fine-tuned on real observations;
    \item trained on virtual observations, fine-tuned on virtual observations adapted with CycleGAN~\cite{cyclegan} and then further fine-tuned on real observations.
\end{enumerate}

CycleGAN has been considered in the experiments as a form of unsupervised domain adaptation to assess if it is beneficial to adapt the virtual observations to the real-world images.
We trained CycleGAN~\cite{cyclegan} for 50 epochs using two sets of images: a set of 5k images randomly sampled from the virtual 3D model through Habitat~\cite{habitat19iccv} and another set of 5k images randomly sampled from the train set of the real-world images.

\subsection{Navigation Episodes with Sensors and Actuation Nois\-es}
We considered three increasing levels of localization noise and three increasing levels of actuation noises, small, medium, and large, as reported in Table \ref{table:noise_levels}. 
The medium actuation noise values have been estimated from measurements collected performing actions such as ``go straight by $0.25m$'' and ``rotate by 10$^\circ$'' on the considered Sanbot Elf robotic platform.
The medium localization noise values have been chosen considering the typical error of an image-based indoor localization system~\cite{arewemakingprogress}. 

We evaluated the impact of localization and sensor noise by training visual navigation models in simulation with and without noise and testing them with progressive levels of noise. 
Specifically, we first trained the navigation models without noise and then fine-tuned them with small, medium and large sensor and actuator noise levels and with different combinations of them, after freezing the visual encoder weights. 

\subsection{Evaluation}
We evaluated the performance of the visual navigation models in terms of SPL and success rate in reaching the goal in a limited number of steps, which are two standard evaluation measure for the visual navigation task. The SPL is a measure of the efficiency of the navigation episode with respect to the shortest geodesic path and is defined as:
\begin{equation}
    \frac{1}{N}\sum_{i=1}^{N}{S_{i}\frac{l_{i}}{max(l_{i}, p_{i})} }
\end{equation}
where $N$ is the number of performed episodes, $S_i$ is a boolean indicator of the success of the \textit{i-th} episode, $l_i$ is the shortest geodesic path length from the starting position to the goal position of the \textit{i-th} episode and $p_i$ is the agent's path length in the \textit{i-th} episode. If the navigation episode follows exactly the shortest path, it assumes the value of 1. On the contrary if the navigation policy fails, it assumes the value of 0.
We randomly sampled 1000 navigation episodes defined by a starting and a goal positions, thus we stored them and used them to evaluate all the navigation models. To avoid to sample excessively simple navigation episodes, the navigation episodes have been filtered to ensure that the ratio between the geodesic distance and the euclidean distance from the starting position to the goal is greater than $1.1$, as already suggested in~\cite{habitat19iccv}.
We set the episodic maximum number of steps threshold to 200. We believe this is a reasonable value given the size and the complexity of the environment. 
An episode is considered successful if the agent calls the \textit{STOP} action within $0.20m$ from the goal and unsuccessful otherwise.

\begin{table}[t]
\caption{Virtual to Real Policy Transfer}
\scalebox{0.97}{
\begin{tabular}{ccccc}
\hline
Training Stages       & SPL    & Success rate & \begin{tabular}[c]{@{}c@{}}Avg. dist.\\ from goal\\ (meters)\end{tabular} & \begin{tabular}[c]{@{}c@{}}\# training\\ frames\\ (Million)\end{tabular} \\ \hline
\rowcolor[HTML]{D9D9D9} 
Virtual               & 0.0160 & 0.022        & 7.9722                                                                    & 2.4                                                                     \\
CycleGAN              & 0.2464 & 0.3310       & 4.6065                                                                    & 2.4                                                                     \\
\rowcolor[HTML]{D9D9D9} 
Virtual+CycleGAN      & 0.2648 & 0.3410       & 4.7535                                                                    & 1.2+1.2                                                                 \\
Real                  & 0.7112 & 0.8590       & 0.7709                                                                    & 2.4                                                                     \\
\rowcolor[HTML]{D9D9D9} 
Virtual+real          & 0.8001 & 0.9700       & 0.2493                                                                    & 1.2+1.2                                                                 \\
CycleGAN+real         & 0.7665 & 0.8880       & 0.5219                                                                    & 1.2+1.2                                                                 \\
\rowcolor[HTML]{D9D9D9} 
Virtual+CycleGAN+real & 0.7553 & 0.9360       & 0.3313                                                                    & 1.2+1.2+1.2                                                             \\ \hline
\end{tabular}}
\label{table:results_real}
\end{table}

\section{Results}

\label{sec:results}

\subsection{Virtual to Real Policy Transfer}
Table~\ref{table:results_real} reports the results of the compared methods when trained on different combinations of virtual and real observations in the absence of sensor and actuator noise, and tested on real-world observations. 
For each experiment, we report the set of training stages used to learn the policy. The notation $A+B$ indicates that the model has been first trained on $A$, then fine-tuned on $B$. ``Virtual'' and  ``Real'' denote that the model has been trained with virtual and real-world observations respectively, whereas ``CycleGAN'' indicates that the model has been trained on virtual observations translated to real using CycleGAN. The last column of the table reports the number of training frames seen in each stage.
As can be seen, the approach trained only with virtual observations achieves very limited performance, e.g., ``Virtual'' obtains an SPL of $0.0160$ and a success rate of $0.0220$. Performance improves when CycleGAN is used to reduce the domain gap. Indeed, ``CycleGAN'' obtains a better success rate of $0.3310$.
Pre-training the model with $1.2M$ frames on virtual observations, and then fine-tuning with observations translated with CycleGAN allows for minor improvements on all measures. For instance, the $SPL$ of ``Virtual + CycleGAN'' is equal to $0.2648$, which is better than ``CycleGAN'' by about two percentage points.
Since the total number of frames seen during training by the two approaches is $2.4M$ in both cases, we speculate that the minor improvement might be due to the virtual observations being less noisy than the ones translated with CycleGAN, which might improve generalization ability. It is worth noting that, while these improvements might seem scarce, the approaches using CycleGAN do not rely on any real-world labels such as the camera poses obtained using structure from motion. As such, they suggest a promising research direction for virtual to real policy transfer.

Interestingly, training the agent with real observations using the proposed tool allows to obtain major improvements in performance. For instance, ``Real'' achieves a success rate of $0.8590$, which is a significant improvement over the success rate of $0.3410$ obtained by ``Virtual+CycleGAN''. Pre-training the model with virtual observations allows to obtain additional improvements bringing the SPL to $0.8001$ (``Virtual + real''), the success rate to $0.97$ and the average distance from the goal to $0.2493$ meters. This result suggests that the knowledge learned in a purely simulated environment can be transferred to real-world observations using appropriate tools. 
Combining CycleGAN with training on real-world observations does not lead to improvements. We believe this is due to the limited ability of CycleGAN to translate virtual observations to real.

\begin{figure}[t]
    \centering
    \includegraphics[width=0.99\linewidth]{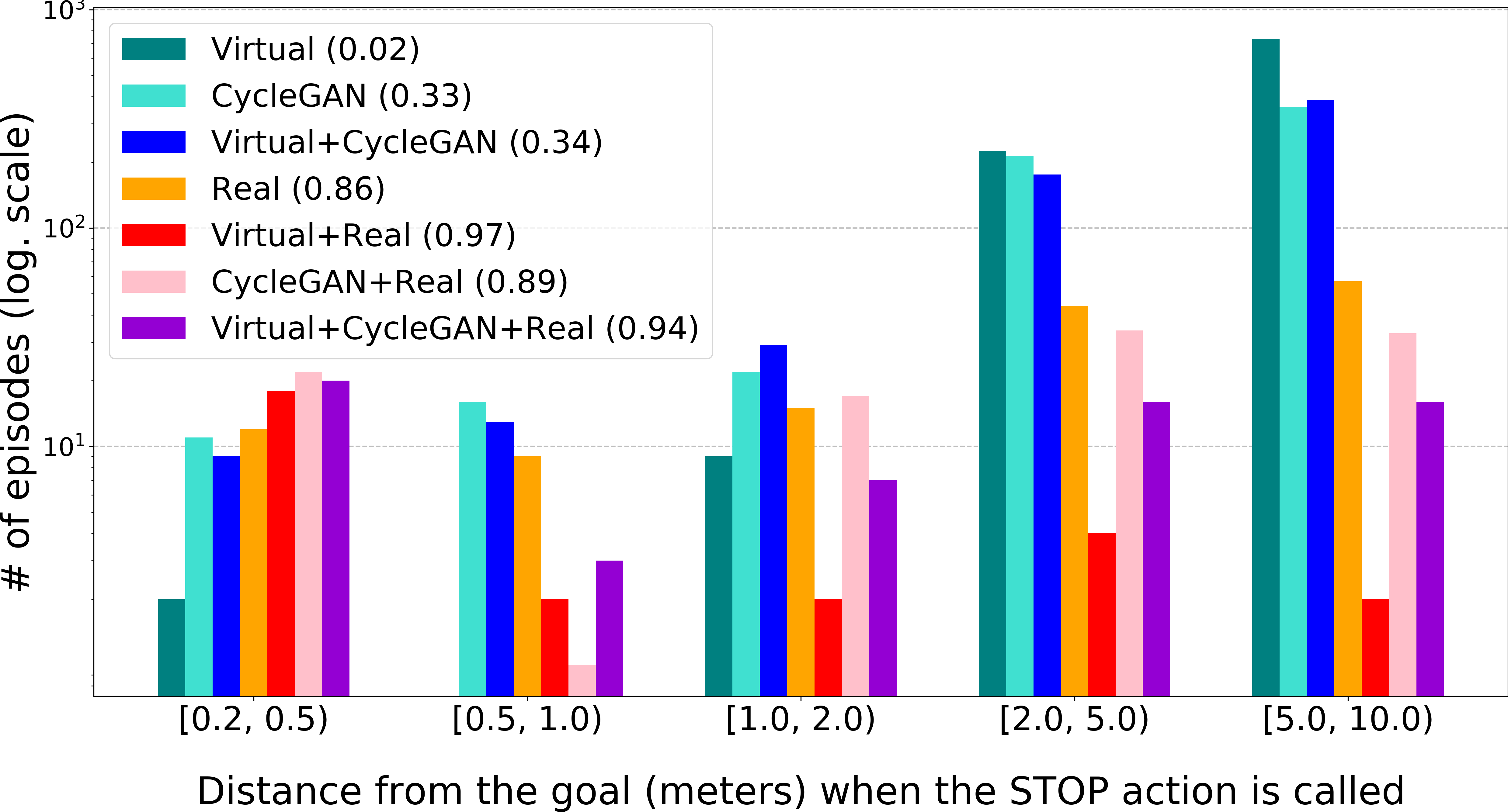}
    \caption{Distribution of unsuccessful navigation episodes with respect to the final distance from the goal for models evaluated on real-world observations. Success rates are reported in parenthesis in the legend.}
    \label{figure: distance_from_goal_real}
\end{figure}

Figure~\ref{figure: distance_from_goal_real} reports the distribution of the unsuccessful navigation episodes with respect to the final distance from the goal for all the methods discussed in this section. Specifically, each bin accumulates the number of episodes for which the distance of the agent from the goal when the STOP action has been called was in a given range. Since episodes are considered unsuccessful when this distance is above $0.2m$, the plot aims to assess qualitatively how far the agent is from the goal when it fails. As shown in the plot, in most of the unsuccessful episodes, the agent trained only on virtual observations (``Virtual'') tend to be located very far from the target with a peak in the range $[5.0, 10.0)$. Training with adapted visual observations (``CycleGAN'' and ``Virtual+CycleGAN'') allows to obtain a flatter distribution, with many failures moving to the range $[0.2, 0.5)$. When the model is trained  with real observations, the failures are more evenly distributed across bins. Interestingly, ``Virtual+Real'' exhibits a distribution skewed towards the left side, which indicates that most of the unsuccessful episodes have been terminated in the proximity of the target. Combining CycleGAN with real data (``CycleGAN+Real'' and ``Virtual+CycleGAN+Real'') does not bring significant benefits as already observed in Table~\ref{table:results_real}.
\begin{table}[t]
\caption{Sensor and actuator noise with virtual observations}
\centering
\begin{tabular}{cccccc}
\hline
\rule{0pt}{5ex}
\begin{tabular}[c]{@{}c@{}}Sensors\\ noise\end{tabular} & \begin{tabular}[c]{@{}c@{}}Actuators\\ noise\end{tabular} & \begin{tabular}[c]{@{}c@{}}Trained\\ with\\ noise\end{tabular} & SPL    & \begin{tabular}[c]{@{}c@{}}Success\\ rate\end{tabular} & \begin{tabular}[c]{@{}c@{}}Avg. dist. \\ from goal\\ (meters)\end{tabular} \\[1ex]
\hline
\rule{0pt}{3ex}
No                                                      & No                                                        & No                                                             & 0.9127 & 0.9910                                                 & 0.1291                                                                     \\[1ex]
\rowcolor[HTML]{D9D9D9} 
\cellcolor[HTML]{D9D9D9}                                & \cellcolor[HTML]{D9D9D9}                                  & No                                                             & 0.8173 & 0.8910                                                 & 0.1581                                                                     \\
\rowcolor[HTML]{D9D9D9} 
\multirow{-2}{*}{\cellcolor[HTML]{D9D9D9}Small}         & \multirow{-2}{*}{\cellcolor[HTML]{D9D9D9}No}              & Yes                                                             & 0.8658 & 0.9380                                                 & 0.1065                                                                     \\
                                                        &                                                           & No                                                             & 0.5075 & 0.5660                                                 & 0.2404                                                                     \\
\multirow{-2}{*}{Medium}                                & \multirow{-2}{*}{No}                                      & Yes                                                             & 0.7114 & 0.7910                                                 & 0.1554                                                                     \\
\rowcolor[HTML]{D9D9D9} 
\cellcolor[HTML]{D9D9D9}                                & \cellcolor[HTML]{D9D9D9}                                  & No                                                             & 0.1552 & 0.1870                                                 & 0.4909                                                                     \\
\rowcolor[HTML]{D9D9D9} 
\multirow{-2}{*}{\cellcolor[HTML]{D9D9D9}Large}         & \multirow{-2}{*}{\cellcolor[HTML]{D9D9D9}No}              & Yes                                                             & 0.3643 & 0.4130                                                 & 0.2577                                                                     \\
                                                        &                                                           & No                                                             & 0.9092 & 0.9890                                                 & 0.1171                                                                     \\
\multirow{-2}{*}{No}                                    & \multirow{-2}{*}{Small}                                   & Yes                                                             & 0.9073 & 0.9820                                                 & 0.0956                                                                     \\
\rowcolor[HTML]{D9D9D9} 
\cellcolor[HTML]{D9D9D9}                                & \cellcolor[HTML]{D9D9D9}                                  & No                                                             & 0.8903 & 0.9700                                                 & 0.1432                                                                     \\
\rowcolor[HTML]{D9D9D9} 
\multirow{-2}{*}{\cellcolor[HTML]{D9D9D9}No}            & \multirow{-2}{*}{\cellcolor[HTML]{D9D9D9}Medium}          & Yes                                                             & 0.8805 & 0.9740                                                 & 0.1150                                                                     \\
                                                        &                                                           & No                                                             & 0.8043 & 0.8860                                                 & 0.2337                                                                     \\
\multirow{-2}{*}{No}                                    & \multirow{-2}{*}{Large}                                   & Yes                                                             & 0.8381 & 0.9340                                                 & 0.2322                                                                     \\
\rowcolor[HTML]{D9D9D9} 
\cellcolor[HTML]{D9D9D9}                                & \cellcolor[HTML]{D9D9D9}                                  & No                                                             & 0.8020 & 0.8740                                                 & 0.1876                                                                     \\
\rowcolor[HTML]{D9D9D9} 
\multirow{-2}{*}{\cellcolor[HTML]{D9D9D9}Small}         & \multirow{-2}{*}{\cellcolor[HTML]{D9D9D9}Small}           & Yes                                                             & 0.8328 & 0.8950                                                 & 0.1607                                                                     \\
                                                        &                                                           & No                                                             & 0.4537 & 0.5100                                                 & 0.2675                                                                     \\
\multirow{-2}{*}{Medium}                                & \multirow{-2}{*}{Medium}                                  & Yes                                                             & 0.4715 & 0.5290                                                 & 0.2712                                                                     \\
\rowcolor[HTML]{D9D9D9} 
\cellcolor[HTML]{D9D9D9}                                & \cellcolor[HTML]{D9D9D9}                                  & No                                                             & 0.1288 & 0.1620                                                 & 0.5442                                                                     \\
\rowcolor[HTML]{D9D9D9} 
\multirow{-2}{*}{\cellcolor[HTML]{D9D9D9}Large}         & \multirow{-2}{*}{\cellcolor[HTML]{D9D9D9}Large}           & Yes                                                             & 0.2450 & 0.2790                                                 & 0.4040                                                                    
\\
\hline                                                             
\end{tabular}
\label{table: w/_w/o_noise}
\end{table}

\subsection{Influence of actuator and sensor noise}
Table~\ref{table: w/_w/o_noise} reports the performance of navigation models when trained and tested in the presence of different combinations of sensor and actuator noise. This analysis is performed on virtual observations as these provide a much more accurate geometrical representation of the space.
We consider two main approaches: 1) training the model without noise and testing it with noise, 2) training and testing the model with the same amount of noise.
The visual navigation model trained and tested on virtual images in the absence of noise obtains good SPL and success rate values of $0.9127$ and $0.9910$ respectively (first row ``No-No-No'' in Table~\ref{table: w/_w/o_noise}).
This suggests that such methods can successfully learn near-optimal navigation policies when trained and tested in simulation. 
However, even adding small amounts of sensor or actuation noise during test degrades performance. For instance an SPL of $0.8658$ is achieved when the model is tested with small sensor noise (``Small-No-No'' in Table~\ref{table: w/_w/o_noise}) and an SPL of $0.9073$ when tested with small actuator noise (``No-Small-No'' in Table~\ref{table: w/_w/o_noise}).
The drop in performance is wider as larger amounts of noise and combinations of sensor and actuator noise are considered.
For instance, the model obtains a small success rate of $0.1620$ when tested with large sensor and actuator noise (``Large-Large-No'' in Table~\ref{table: w/_w/o_noise}).
Interestingly, part of the gap in performance is generally recovered by training the methods with the same kind of noise.
For instance, training the model with large amounts of sensor and actuator noise brings the success rate from $0.1620$ to $0.2790$ (``Large-Large-Yes'' in Table~\ref{table: w/_w/o_noise}).
Similarly, training the model with small sensor noise bring the success rate from $0.8910$ to $0.9380$ (compare ``Small-No-No'' to ``Small-No-Yes'' in Table~\ref{table: w/_w/o_noise}).
Similar trends are observed with other combinations of sensor and actuator noise.

\begin{figure}[t]
    \centering
    \includegraphics[width=0.99\linewidth]{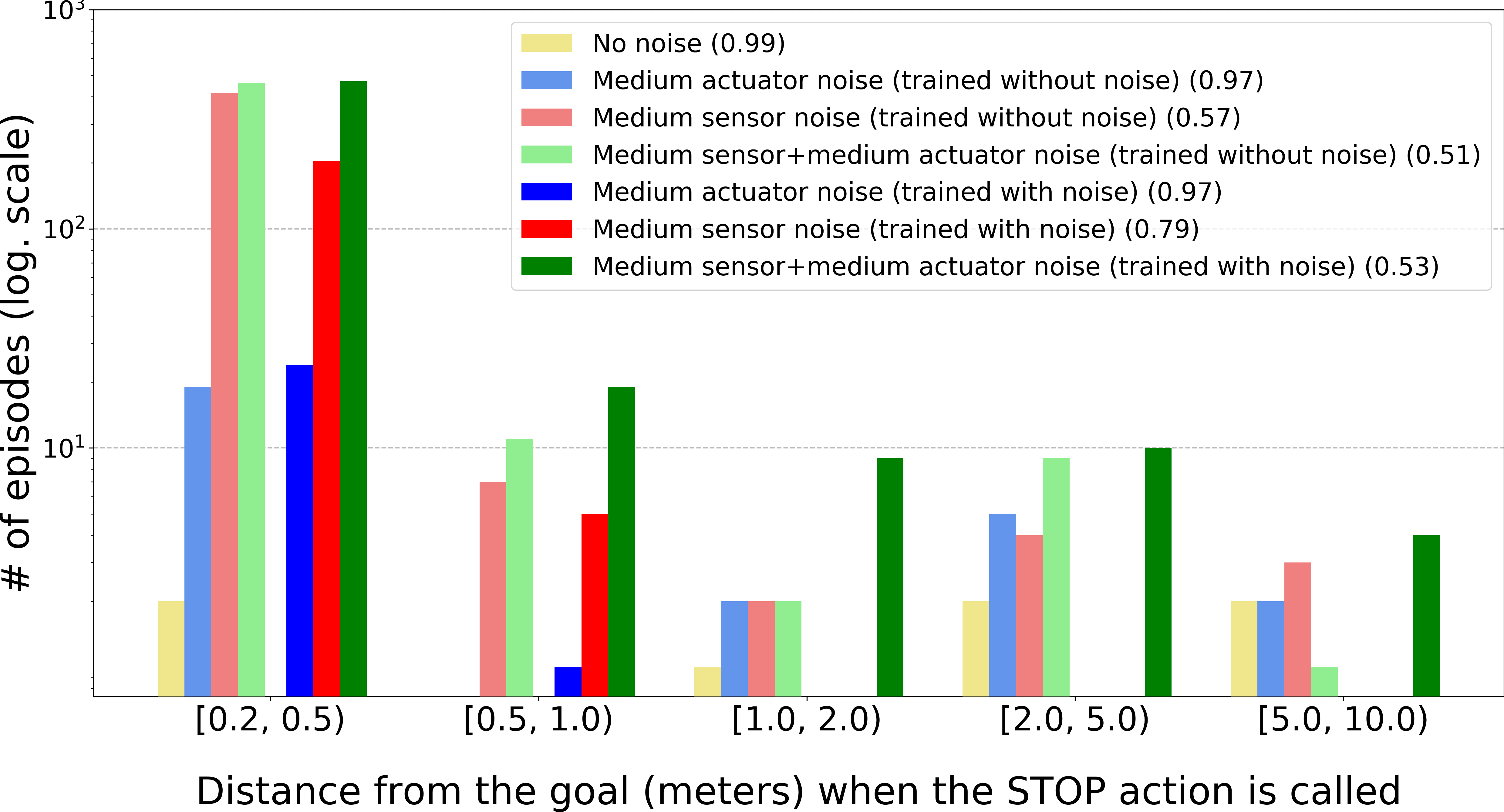}
    \caption{Distribution of unsuccessful navigation episodes with respect to the final distance from the goal for models trained and evaluated on virtual observations with and without sensor and actuator noise. Success rates are reported in parenthesis in the legend.}
    \label{distance_from_goal}
\end{figure}

Figure~\ref{distance_from_goal} reports the distribution of the unsuccessful navigation episodes with respect to the final distance from the goal for some of the experiments considered in this section.
As can be noted, while the distribution of unsuccessful episodes is overall uniform when the model is trained and tested without noise (``No noise'' in Figure~\ref{distance_from_goal}), methods trained without noise but tested with medium actuator or sensor noise tend to terminate unsuccessful episodes at large distances from the target (see ``Medium sensor/actuator noise (trained without noise)'' in Figure~\ref{distance_from_goal}).
The same approaches trained in the presence of noise (see ``Medium sensor/actuator noise (trained with noise)'' in Figure~\ref{distance_from_goal}) are characterized by a distribution which is much more skewed towards small distances. This indicates that most of the episodes terminating at large distances from the target (e.g., at a distance larger than $1m$) have been recovered, and most of the failures are due to the inability of the agent to reach the target in the final part of the navigation episode in which more precision is required.
Different considerations apply to the case of the combination of medium sensor and actuator noise (compare ``Medium sensor + medium actuator noise (trained without noise)'' and ``Medium sensor + medium actuator noise (trained with noise)'' in Figure~\ref{distance_from_goal}).
Even if the success rate improves when the model is trained in the presence of noise, the two distributions do not show noticeable differences.

\subsection{Influence of actuator and sensor noise with real-world observation}
Table~\ref{table: noise_real_images} finally reports the results of training and testing navigation policies on real-world observations in the presence of sensor and actuator noise. 
All models have been trained for $1.2M$ frames on virtual observations without noise, then fine-tuned for $1.2M$ frames on real-world observations with or without noise as specified in the table.
Similarly to what observed in the case of virtual observations, models trained without noise perform worse when tested with noise. Indeed, while the baseline method has an SPL of $0.8001$ (``No-No-No'' in Table~\ref{table: noise_real_images}), the same model obtains an SPL of $0.6553$ when trained with small sensor and actuator noise (``Small-Small-No'' in Table~\ref{table: noise_real_images}), $0.4041$ when trained with medium sensor and actuator noise (``Medium-Medium-No'' in Table~\ref{table: noise_real_images}), and $0.1467$ when trained with large sensor and actuator noise (``Medium-Medium-No'' in Table~\ref{table: noise_real_images}).
This confirms that noise plays an important role in the quality of the execution of navigation policies.
Differently from what observed for virtual observations, training the same models with noise leads to even worse results (compare the performance of models trained with and without noise in Table~\ref{table: noise_real_images}).
We speculate that this might be due to the more challenging nature of the navigation task with real-world observations and suggest that more sophisticated techniques to compensate for noise should be investigated.

\begin{table}[t]
\caption{Sensor and actuator noise with real-world observations}
\centering
\begin{tabular}{cccccc}
\hline
\rule{0pt}{5ex}
\begin{tabular}[c]{@{}c@{}}Sensors\\ noise\end{tabular} & \begin{tabular}[c]{@{}c@{}}Actuators\\ noise\end{tabular} & \begin{tabular}[c]{@{}c@{}}Trained\\ with\\ noise\end{tabular} & SPL    & \begin{tabular}[c]{@{}c@{}}Success\\ rate\end{tabular} & \begin{tabular}[c]{@{}c@{}}Avg. dist. \\ from goal\\ (meters)\end{tabular} \\ [1ex]
\hline
\rule{0pt}{3ex}
No                                                      & No                                                        & No                                                             & 0.8001 & 0.9700                                                 & 0.2493                                                                     \\[1ex]
\rowcolor[HTML]{D9D9D9} 
\cellcolor[HTML]{D9D9D9}                                & \cellcolor[HTML]{D9D9D9}                                  & No                                                             & 0.6553 & 0.7880                                                 & 0.6564                                                                     \\
\rowcolor[HTML]{D9D9D9} 
\multirow{-2}{*}{\cellcolor[HTML]{D9D9D9}Small}         & \multirow{-2}{*}{\cellcolor[HTML]{D9D9D9}Small}           & Yes                                                             & 0.4708 & 0.6270                                                 & 0.7269                                                                     \\
                                                        &                                                           & No                                                             & 0.4041 & 0.4041                                                 & 0.7771                                                                     \\
\multirow{-2}{*}{Medium}                                & \multirow{-2}{*}{Medium}                                  & Yes                                                             & 0.3116 & 0.4150                                                 & 0.8231                                                                     \\
\rowcolor[HTML]{D9D9D9} 
\cellcolor[HTML]{D9D9D9}                                & \cellcolor[HTML]{D9D9D9}                                  & No                                                             & 0.1467 & 0.1870                                                 & 0.9808                                                                     \\
\rowcolor[HTML]{D9D9D9} 
\multirow{-2}{*}{\cellcolor[HTML]{D9D9D9}Large}         & \multirow{-2}{*}{\cellcolor[HTML]{D9D9D9}Large}           & Yes                                                             & 0.0827 & 0.1180                                                 & 1.9109
\\
\hline
\end{tabular}
\label{table: noise_real_images}
\end{table}

\section{Conclusion}

\label{sec:conclusion}

We investigated the problem of transferring visual navigation policies trained in simulation to the real world.
We proposed a tool based on Habitat~\cite{habitat19iccv} which can be used to produce virtual and real-world navigation episodes accounting for sensor and actuator noise models.
Using the tool, we assessed the impact of the domain shift between virtual and real-world navigation induced by the difference in appearance and physical interactions.
Our experiments suggest that adaptation methods are much needed to obtain visual navigation policies able to generalize to the real world.
We think that the proposed framework is a promising tool to assess and improve the generalization ability of navigation policies to the real world and hope that it will encourage research in this direction.

\section*{Acknowledgment}
This research is supported by OrangeDev s.r.l., by  Piano della Ricerca 2016-2018 - CHANCE - Linea di Intervento 1 of DMI, University of Catania, and by MIUR AIM - Attrazione e Mobilit\`a Internazionale Linea 1 - AIM1893589 - CUP E64118002540007. 



\bibliographystyle{ieeetr}
\bibliography{bibliography}
%

\end{document}